%% file: main.tex
\documentclass[letterpaper]{article} 
\usepackage{aaai24}  
\usepackage{times}  
\usepackage{helvet}  
\usepackage{courier}  
\usepackage[hyphens]{url}  
\usepackage{graphicx} 
\usepackage{natbib}  
\usepackage{caption} 
\usepackage{algorithm}
\usepackage{algorithmic}

\usepackage{newfloat}
\usepackage{listings}
\DeclareCaptionStyle{ruled}{labelfont=normalfont,labelsep=colon,strut=off} 
\floatstyle{ruled}
\newfloat{listing}{tb}{lst}{}
\floatname{listing}{Listing}
\title{Understanding CNN Hidden Neuron Activations Using Structured Background Knowledge and Deductive Reasoning}
\author {
    Abhilekha Dalal,\textsuperscript{\rm 1}
    Md Kamruzzaman Sarker,\textsuperscript{\rm 2}
    Adrita Barua,\textsuperscript{\rm 1}
    Eugene Vasserman,\textsuperscript{\rm 1}
    Pascal Hitzler\textsuperscript{\rm 1}
}
\affiliations {
    \textsuperscript{\rm 1} Kansas State University\hspace{1cm}
    \textsuperscript{\rm 2} Bowie State University \\
    adalal@ksu.edu, mdkamruzzamansarker@gmail.com, adrita@ksu.edu, eyv@ksu.edu, hitzler@ksu.edu
}

\usepackage{bibentry}

\begin{document}

\maketitle

\begin{abstract}
A major challenge in Explainable AI is in correctly interpreting activations of hidden neurons: accurate interpretations would provide insights into the question of what a deep learning system has internally \emph{detected} as relevant on the input, demystifying the otherwise black-box character of deep learning systems.
The state of the art indicates that hidden node activations can, in some cases, be interpretable in a way that makes sense to humans, but systematic automated methods that would be able to hypothesize and verify interpretations of hidden neuron activations are underexplored.

In this paper, we provide such a method and demonstrate that it provides meaningful interpretations. Our approach is based on using large-scale background knowledge --
approximately 2 million classes curated from the Wikipedia concept hierarchy -- together with a symbolic reasoning approach called \emph{Concept Induction} based on description logics, originally developed for applications in the Semantic Web field.
Our results show that we can automatically attach meaningful labels from the background knowledge to individual neurons in the dense layer of a Convolutional Neural Network through a hypothesis and verification process. 
\end{abstract}

\section{Introduction}
\label{introduction}
Deep learning has led to significant advances in artificial intelligence applications including image classification~\cite {ramprasath2018image}, speech recognition~\cite{graves2014towards}, translation~\cite{auli2013joint}, drug design~\cite{segler2018generating}, medical diagnosis~\cite{choi2019artificial}, climate sciences~\cite{liu2016application}, and many more.
Despite these successes, the black-box nature of deep learning systems remains problematic for some application areas, especially those involving automated decisions and safety-critical systems. For example, Apple co-founder Steve Wozniak accused Apple of gender discrimination, claiming that the new Apple Card gave him a credit limit that was ten times higher than that of his wife even though the couple shares all property~\cite{apple_card}.
In an image search, only 11\% of the top image results for ``CEOs'' were images of women despite the fact that women make up 27\% of US CEOs~\cite{CEO_bias}. Other application areas
of particular concern
include safety-critical systems such as self-driving cars~\cite{chen2017end}, drug discovery and treatment recommendations~\cite{rifaioglu2020deepscreen,hariri2021deep}, and others, as deep learning systems are prone to adversarial attacks, e.g., by altering classification results by introducing adversarial examples~\cite{bau2020understanding} or simply controlling the order in which training images are presented~\cite{data_order}. Some of these attacks are difficult or impossible to detect after the fact~\cite{undetectable_backdoor,compile_backdoor}.

Standard assessments of deep learning
performance consist of statistical evaluation, but do not seem sufficient to address these shortcomings as they cannot provide reasons or explanations for particular system behaviors~\cite{doran2018does}.
Consequently, it remains very important to develop strong explanation methods for deep learning systems. While there has been significant progress on this front (see Section~\ref{relatedWork}), the current state of the art is mostly restricted to explanation analyses based on a relatively small number of predefined explanation categories. This is problematic from a principled perspective, as this relies on the assumption that explanation categories pre-selected by humans would be viable explanation categories for deep learning systems -- an as-yet unfounded conjecture. Other state of the art explanation systems rely on modified deep learning architectures, usually leading to a decrease in system performance compared to unmodified systems~\cite{DBLP:conf/nips/ZarlengaBCMGDSP22}.
Ideally, we would want strong explanation capabilities while maintaining the underlying learning architecture.

In this paper, we address the aforementioned shortcomings by using
\emph{Concept Induction}, i.e., formal logical deductive reasoning~\cite{DBLP:journals/ml/LehmannH10}. We show that our approach can indeed provide meaningful explanations for hidden neuron activation in a
Convolutional Neural Network (CNN) architecture for image scene classification (on the ADE20K dataset~\cite{zhou2019semantic}), using a class hierarchy consisting of about $2 \cdot 10^6$ classes, derived from Wikipedia, as the pool of categories~\cite{DBLP:conf/kgswc/SarkerSHZNMJRA20}. The benefits of our approach to explainable deep learning are: (a) it can be used on unmodified and pre-trained deep learning architectures, (b) it assigns semantic categories (i.e., class labels expressed in formal logic) to hidden neurons such that images related to these labels activate the corresponding neuron with high probability, and (c) it can construct these labels from a very large pool of categories.

The rest of this paper is organized as follows. Section~\ref{relatedWork} discusses related work on explainable deep learning. Section~\ref{researchMethod} presents our approach. Section~\ref{sec:eval} provides evaluation results and Section~\ref{sec:discussion} discussions thereof. Section~\ref{conclusion} concludes and discusses follow-up research directions. A technical appendix provides more complete details of our experiments and results. 
Source code, input data, raw result files, and parameter settings for replication are available online.\footnote{\url{https://github.com/abhilekha-dalal/xai-using-wikidataAndEcii/}}

\section{Related Work}
\label{relatedWork}
Explaining
(interpreting, understanding, justifying) automated AI decisions has been explored from the early 1970s. With the recent advances in deep learning~\cite{lecun2015deep}, its wide usage in nearly every field, and its opaque nature make explainable AI more important than ever,
and there are multiple ongoing efforts to demystify deep learning~\cite{gunning2019xai, adadi2018peeking, minh2022explainable}. 
Existing explainable methods can be categorized based on input data (feature) understanding, e.g., feature summarizing~\cite{selvaraju2016grad, ribeiro2016should}, or based on the model's internal unit representation, e.g., node summarizing~\cite{zhou2018interpreting,bau2020understanding}. Those methods can be further categorized as model-specific~\cite{selvaraju2016grad} or model-agnostic~\cite{ribeiro2016should}. 
Another kind of approach relies on human interpretation of explanatory data returned, such as counterfactual questions~\cite{DBLP:journals/corr/abs-1711-00399}. 

We focus on the understanding of internal units of the neural network-based deep learning models. Given a deep learning model and its prediction, we ask the questions ``What does the deep learning model's internal unit represent? Are those units activated by human-understandable concepts?''
Prior work has shown that internal units may indeed represent human-understandable concepts~\cite{zhou2018interpreting,bau2020understanding}, but these approaches require semantic segmentation~\cite{xiao2018unified}  (which is time- and compute-expensive) or explicit concept annotations~\cite{kim2018interpretability} (which are expensive to acquire).
To get around these limitations, we
take a different approach
by using
a hypothesis generation and validation approach based on Concept Induction analysis for hypothesis generation (details in Section~\ref{sec:approach}).
The use of large-scale
description logic background knowledge
means that we draw explanations from a very large pool of explanation categories.

There has been some work using knowledge graphs to produce explanations from  deep learning models~\cite{confalonieri2021using, diaz2022explainable}, and also on using Concept Induction to provide explanations~\cite{sarker2017explaining, 9736291}, but they focused on analysis of input-output behavior, i.e., on generating an explanation for the overall system. We focus instead on the different task of understanding internal (hidden) node activations. 

To the best of our knowledge, our use of Concept Induction with large-scale background knowledge as pool for explanation generation (to understand the internal node activations) is novel. 
Furthermore our method (training, Concept Induction analysis, and verification) is fully automateable without the need for human intervention.

\section{Approach}
\label{researchMethod}\label{approach}\label{sec:approach}

In this section we detail our technical approach. Section~\ref{subsec:training} covers the
scene recognition scenario that we use to present our approach; Section~\ref{subsec:CI}
describes the technical components used for explanation generation; Section~\ref{subsec:label-hypotheses} presents our results and how we obtain label hypotheses
for hidden node activations (with examples); and Section~\ref{subsec:label-validation} details
the label hypothesis validation process and results. Experimental evaluation can be found in
Section~\ref{resultsDiscussion}. More details regarding our experimental
parameters are in Appendix~\ref{appendix}.

\subsection{Preparations: Scenario and CNN Training}
\label{subsec:training}

We use a scene classification from images scenario
to demonstrate our approach, drawing from the ADE20K dataset~\cite{zhou2019semantic} which
contains more than 27,000 images over 365 scenes, extensively annotated with pixel-level
objects and object part labels. \emph{The annotations are not used for CNN training}, but rather only for generating label hypotheses that we will describe in Section~\ref{subsec:label-hypotheses}.

\begin{table}[tb]
\centering
\begin{tabular}{lcc}
    Architectures  & Training acc & Validation acc \\
    \hline
    Vgg16          & 80.05\%      & 46.22\%        \\
    InceptionV3    & 89.02\%      & 51.43\%         \\
    Resnet50       & 35.01\%      & 26.56\%        \\
\textbf{Resnet50V2}    &\textbf{87.60\%}      &\textbf{86.46\%}  \\
    Resnet101      & 53.97\%      & 53.57\%        \\
    Resnet152V2    & 94.53\%      & 51.04\%        \\
\end{tabular}
\caption{Performance (accuracy) of different architectures on the ADE20K dataset.
The system we used, based on performance, is bolded.}
\label{accuracy_of_networks}
\end{table}

We train a classifier for the following scene
categories: ``bathroom,'' ``bedroom,'' ``building facade,'' ``conference room,'' ``dining
room,'' ``highway,'' ``kitchen,'' ``living room,'' ``skyscraper,'' and ``street.'' We weigh our selection toward scene
categories which have the highest number of images and we deliberately include some scene
categories that should have overlapping annotated objects -- we believe this makes the
hidden node activation analysis more interesting. We did not conduct any experiments on any
other scene selections yet, i.e., \emph{we did not change our scene selections based on
any preliminary analyses}.

We trained a number of CNN architectures in order to use the one with highest accuracy, namely Vgg16~\cite{simonyan2015very}, InceptionV3~\cite{szegedy2016rethinking} and different versions of Resnet -- Resnet50, Resnet50V2, Resnet101, Resnet152V2~\cite{he2016deep,he2016identity}.
Each neural network was fine-tuned with a dataset of 6,187 images (training and validation
set) of size 224x224 for 30 epochs with early stopping\footnote{monitoring validation loss; patience 3; restoring best weights}
to avoid overfitting.
We used Adam as our optimization algorithm, with a categorical cross-entropy loss function and a learning rate of 0.001.

We select Resnet50V2 because it achieves the highest accuracy (see Table~\ref{accuracy_of_networks}). Note that for our investigations, which focus on explainability of hidden neuron activations, achieving a very high accuracy for the scene classification task is not essential, but a reasonably high accuracy
is necessary when considering models which would be
useful
in practice.

\subsection{Preparations: Concept Induction and Background Knowledge}
\label{subsec:CI}

For label hypotheses generation, we make use of \emph{Concept Induction}~\cite{DBLP:journals/ml/LehmannH10}
which is based on deductive reasoning over description logics, i.e., over logics relevant to ontologies, knowledge graphs, and generally the Semantic Web field~\cite{DBLP:books/crc/Hitzler2010,DBLP:journals/cacm/Hitzler21}.
Concept Induction has indeed already been shown, in other scenarios, to be capable of producing labels that are meaningful for humans inspecting the data~\cite{DBLP:journals/corr/abs-2209-13710}.
A Concept Induction system accepts three inputs: (1) a set of positive examples $P$, (2) a set of
negative examples $N$, and (3) a knowledge base (or ontology) $K$, all expressed as description
logic theories, 
and all examples $x\in P\cup N$ occur as instances (constants) in $K$.
It
returns description logic class expressions $E$ such that $K\models E(p)$ for all $p \in P$ and $K \not\models E(q)$ for all $q \in N$. If no such class expressions exist, then  it returns approximations for $E$ together with a number of accuracy measures.

For scalability reasons, we use the heuristic Concept Induction system
ECII~\cite{DBLP:conf/aaai/SarkerH19} together with a background knowledge base that consists
only of a hierarchy of approximately 2~million classes, curated from the Wikipedia
concept hierarchy and presented in~\cite{DBLP:conf/kgswc/SarkerSHZNMJRA20}. We use
\emph{coverage} as accuracy measure, defined as
$$\textrm{coverage}(E) = \frac{|Z_1 | + |Z_2|}{|P\cup N|},$$
where $Z_1=\{p\in P\mid K\models E(p)\}$ and $Z_2 = \{n\in N\mid K\not\models E(n)\}$.
$P$ is the set of all positive instances, $N$ is the set of all negative instances, and $K$ is the knowledge base provided to ECII as part of the input.

For the Concept Induction analysis, positive and negative example sets will contain images
from ADE20K, i.e., we need to include the images in the background knowledge by linking them
to the class hierarchy. For this, we use the object annotations available for the ADE20K
images, but only part of the annotations for the sake of simplicity. More precisely, we only use the information that certain objects (such as windows) occur in certain images, and we do not make use of any of the richer annotations
such as those related to segmentation. All objects from all images are then mapped to classes in the class hierarchy using the Levenshtein string similarity metric~\cite{DBLP:journals/iandc/Levenshtein75} with edit distance $0$. For example, the ADE20K image \texttt{ADE\_train\_00001556.jpg} has ``door'' listed as one of the objects shown, which is mapped to the ``door'' concept of the Wikipedia concept hierarchy.
Note that the scene information is not used for the mapping, i.e., the images themselves are not assigned to specific (scene) classes in the class hierarchy -- they are connected to the hierarchy only through the objects that are shown (and annotated) in each image.

\subsection{Generating Label Hypotheses}
\label{subsec:label-hypotheses}

The general idea for generating label hypotheses using Concept Induction is as follows: given a hidden neuron,
$P$ is a set of inputs (i.e., in this case, images) to the deep learning system that activate the neuron, and $N$ is a set of inputs that do not activate the neuron
(where $P$ and $N$ are the sets of positive and negative examples, respectively).
As mentioned above, inputs are annotated with classes from the background knowledge for Concept Induction, but these annotations and the background knowledge are not part of the input to the deep learning system. ECII generates a label hypothesis for the given neuron on inputs $P$, $N$, and the background knowledge.

We first feed 1,370 ADE20K images to our trained Resnet50V2 and retrieve the activations of the dense layer.
We chose to look at the dense layer because previous studies indicate~\cite{distill} that earlier layers of a CNN respond to low level features such as lines, stripes, textures, colors, while layers near the final layer respond to higher-level features such as face, box, road, etc. The higher-level features align better with the nature of our background knowledge. 

The dense layer consists of 64 neurons. We chose to analyze each of the neurons separately.
We are aware that activation patterns involving more than one neuron may also be informative in the sense that information may be distributed among several neurons, but the analysis of such activation patterns will be part of follow-up work.

For each neuron, we calculate the maximum activation value across all images.
We then take the positive example set $P$ to consist of all images that activate the
neuron with at least 80\% of the maximum activation value, and the negative example set $N$ to
consist of all images that activate the neuron with at most 20\% of the maximum activation
value (or do not activate it at all). The highest scoring response of running ECII on these
sets, together with the background knowledge described in Section~\ref{subsec:CI}, is shown in
Table~\ref{tab:listofConcepts-abbrv} for each neuron, together with the coverage of the ECII
response. For each neuron, we call its corresponding label the \emph{target label}, e.g.,
neuron $0$ has target label ``building.'' Note that some target labels consist of two  concepts, e.g., ``footboard, chain'' for neuron 49 -- this occurs if the corresponding ECII response carries two class expressions joined by a logical conjunction, i.e., in this example ``footboard $\sqcap$ chain'' (as description logic expression) or $\textnormal{footboard}(x) \land \textnormal{chain}(x)$ expressed in first-order predicate logic.

\begin{table*}[t!]
\centering
\begin{footnotesize}
\begin{tabular}{clrrrr}
    Neuron \# & Obtained Label(s) & Images & Coverage & Target \% & Non-Target \% \\
    \hline
    \textbf{0} & \textbf{building} & \textbf{164} & \textbf{0.997} & \textbf{89.024} & \textbf{72.328} \\
    \textbf{1} & \textbf{cross\_walk} & \textbf{186} & \textbf{0.994} & \textbf{88.710} & \textbf{28.923} \\
    \textbf{3} & \textbf{night\_table} & \textbf{157} & \textbf{0.987} & \textbf{90.446} & \textbf{56.714} \\
    6 & dishcloth, toaster & 106 & 0.999 & 16.038 & 39.078 \\
    11 & river\_water & 157 & 0.995 & 31.847 & 22.309 \\
    \hline
    \textbf{16} & \textbf{mountain, bushes} & \textbf{108} & \textbf{0.995} & \textbf{87.037} & \textbf{24.969} \\
    \textbf{18} & \textbf{slope} & \textbf{139} & \textbf{0.983} & \textbf{92.086} & \textbf{69.919} \\
    \textbf{22} & \textbf{skyscraper} & \textbf{156} & \textbf{0.992} & \textbf{99.359} & \textbf{54.893} \\
    26 & skyscraper, river & 112 & 0.995 & 77.679 & 35.489 \\
    \textbf{30} & \textbf{teapot, saucepan} & \textbf{108} & \textbf{0.998} & \textbf{81.481} & \textbf{47.984} \\
    \hline
    40 & sculpture, side\_rail & 119 & 0.995 & 25.210 & 21.224 \\
    \textbf{41} & \textbf{open\_fireplace, coffee\_table} & \textbf{122} & \textbf{0.992} & \textbf{88.525} & \textbf{16.381} \\
    \textbf{43} & \textbf{central\_reservation} & \textbf{157} & \textbf{0.986} & \textbf{95.541} & \textbf{84.973} \\
    46 & casserole & 157 & 0.999 & 45.223 & 36.394 \\
    \textbf{48} & \textbf{road} & \textbf{167} & \textbf{0.984} & \textbf{100.000} & \textbf{73.932} \\
    \hline
    \textbf{49} & \textbf{footboard, chain} & \textbf{126} & \textbf{0.982} & \textbf{88.889} & \textbf{66.702} \\
    \textbf{51} & \textbf{road, car} & \textbf{84} & \textbf{0.999} & \textbf{98.810} & \textbf{48.571} \\
    \textbf{54} & \textbf{skyscraper} & \textbf{156} & \textbf{0.987} & \textbf{98.718} & \textbf{70.432} \\
    58 & plank, casserole & 80 & 0.998 & 3.750 & 3.925 \\
    \textbf{63} & \textbf{edifice, skyscraper} & \textbf{178} & \textbf{0.999} & \textbf{92.135} & \textbf{48.761}\\
 
    \hline
\end{tabular}
\end{footnotesize}
\caption{Selected representative data as discussed throughout the text (the full version is Table~\ref{tab:listofConcepts} in Appendix~\ref{appendix}).
Images: Number of images used per label. Target \%: Percentage of target images activating the neuron above 80\% of its maximum activation. Non-Target \%: The same, but for all other images.
\textbf{Bold} denotes neurons whose labels are considered confirmed.}
\label{tab:listofConcepts-abbrv}
\end{table*}

\begin{figure*}[tb]
\centering
\includegraphics[clip, trim=0in 6.3in 0in 0in, width=0.99\textwidth]{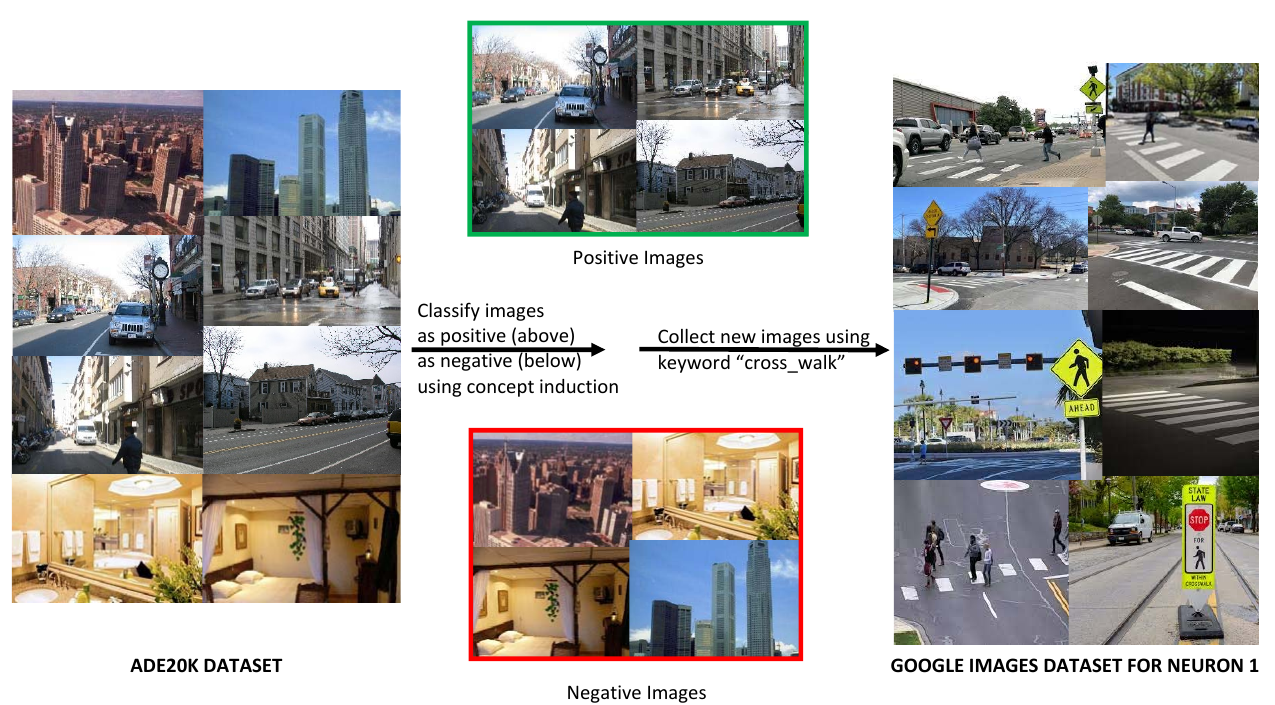} 
\caption{Example of images that were used for generating and confirming the label hypothesis for neuron 1.}
\label{fig1}
\end{figure*}

Let us take neuron 1 as a concrete example.
After training, neuron 1 has
a maximum activation value of 10.90, 80\% of which is 8.72, and 20\% of which is 2.18. The
positive example set $P$ thus consist of all images activating the neuron with at least 8.72,
and the negative example set $N$ consists of all images activating the neuron with at most
2.18. Example images are shown in Figure~\ref{fig1} middle top (positive) and bottom (negative).
The top ranked ECII response on this input was ``cross\_walk,'' with a coverage score of 0.994. (Note that some of the positive images may not actually have a crosswalk, like the top left and bottom right positive images shown in Figure~\ref{fig1} -- we discuss this in Section~\ref{sec:eval}.)
We consider these target labels to be working hypotheses for activation triggers for the corresponding neuron. As hypotheses, they require further confirmation, i.e., some of these hypotheses may be rejected.

\subsection{Confirming Label Hypotheses}
\label{subsec:label-validation}
\label{subsec:label-confirmation}

The process described in Section~\ref{subsec:label-hypotheses} produces label hypotheses for all neurons investigated. The next step is to confirm or reject these hypotheses by testing the labels with new images.\footnote{We would reject labels with low coverage scores, but coverage was $>$ 0.960 across all activated neurons (see Tables~\ref{tab:listofConcepts-abbrv} and~\ref{tab:listofConcepts}).}
We use each of the target labels to search Google Images with the labels as keywords (requiring responses to be returns for \emph{both} keywords if the label is a conjunction of classes). We call each such image a \emph{target image} for the corresponding label or neuron. We use Imageye\footnote{\url{https://chrome.google.com/webstore/detail/image-downloader-imageye/agionbommeaifngbhincahgmoflcikhm}} to automatically retrieve the images, collecting up to 200 images that appear first in the Google Images search results, filtering for images in JPEG format (ADE20K images are in JPEG format) and with a minimum size of 224x224 pixels (again corresponding to ADE20K). For each retrieval label, we use 80\% of the obtained images, reserving the remaining 20\% for the statistical evaluation described in Section~\ref{sec:eval}. The number of images used in the hypothesis confirmation step, for each label, is given in Table~\ref{tab:listofConcepts-abbrv}. These images are fed to the network to check (a) whether the target neuron (with the retrieval label as target label) activates, and (b) whether any other neurons activate.

The Target \% column of Table~\ref{tab:listofConcepts-abbrv} shows the percentage of the target images that activate each neuron to at least 80\% of its maximum activation.
Using neuron 1 as an example, 88.710\% of the images retrieved with the label
``cross\_walk'' activate the neuron ($\ge$ 80\%).
However, this neuron only activates only for 28.923\% of these images,
(indicated in the Non-Target \% column) when presented with images retrieved using all other labels
from Table~\ref{tab:listofConcepts-abbrv} excluding ``cross\_walk.''

We define a target label for a neuron to be \emph{confirmed} if it activates (with at least
80\% of its maximum activation value) for at least 80\% of its target images regardless of
how much or how often it activates for non-target images (see Section~\ref{sec:eval} for the analysis of non-target activation).
We use 80\% as the cut-off for both neuron activation and label hypothesis confirmation --
these are ad-hoc values that could both be chosen differently (we discuss this in  Technical Appendix~\ref{subsubsec:cutoffs}).
This cut-off value
ensures strong association and responsiveness to images retrieved under the target label. We discuss the relevance of the Non-Target column in the next section.

Returning to neuron 1, we retrieve 233 new images with keyword ``cross\_walk,'' 186 of which (80\%)
are used in this step. Example images are shown in Figure~\ref{fig1} to the right. 165 of
these images (i.e., 88.710\%) activate neuron 1 with at least 8.72 (which is 80\% of its
maximum activation value of 10.90). Since $88.710 \geq 80$, we consider the label
``cross\_walk'' confirmed for neuron 1.
After this step, we arrive at a list of
20 \emph{confirmed} labels listed in Table~\ref{tab:evaluation}.

\section{Statistical Evaluation}
\label{resultsDiscussion}
\label{sec:eval}

After generating the confirmed labels (as in Section~\ref{sec:approach}), we statistically
evaluate the node labeling using the remaining images from those retrieved from Google Images
as described in Section~\ref{subsec:label-confirmation}.
Results are shown
in Table~\ref{tab:evaluation}, omitting
 neurons that were not activated
  by any image, i.e., their maximum activation value was 0.
The statistical evaluation shows that Concept Induction analysis with large-scale background knowledge yields meaningful labels that stably explain neuron activation.

\begin{table*}[tb]
\centering
\begin{footnotesize}
\begin{tabular}{clc|rr|rr|rr|rr}
     \multicolumn{1}{l}{Neuron \#} & \multicolumn{1}{l}{Label(s)} & \multicolumn{1}{c}{Images} & \multicolumn{2}{c}{\# Activations (\%)} & \multicolumn{2}{c}{Mean} & \multicolumn{2}{c}{Median} & \multicolumn{1}{c}{z-score} & \multicolumn{1}{c}{p-value}\\
     \hline
     &  & & targ & non-t & targ & non-t & targ & non-t &  & \\
    \hline
    0 & building & 42 & 80.95 & 73.40 & 2.08 & 1.81 & 2.00 & 1.50 & -1.28 & 0.0995 \\
    1 & cross\_walk & 47 & 91.49 & 28.94 & 4.17 & 0.67 & 4.13 & 0.00 & -8.92 & \textless .00001 \\
    3 & night\_table & 40 & 100.00 & 55.71 & 2.52 & 1.05 & 2.50 & 0.35 & -6.84 & \textless .00001 \\
    8 & shower\_stall, cistern & 35 & 100.00 & 54.40 & 5.26 & 1.35 & 5.34 & 0.32 & -8.30 & \textless .00001 \\
    16 & mountain, bushes & 27 & 100.00 & 25.42 & 2.33 & 0.67 & 2.17 & 0.00 & -6.72 & \textless .00001 \\
    \hline
    18 & slope & 35 & 91.43 & 68.85 & 1.59 & 1.37 & 1.44 & 1.00 & -2.03 & 0.0209 \\
    19 & wardrobe, air\_conditioning & 28 & 89.29 & 65.81 & 2.30 & 1.28 & 2.30 & 0.84 & -4.00 & \textless .00001 \\
    22 & skyscraper & 39 & 97.44 & 56.16 & 3.97 & 1.28 & 4.42 & 0.33 & -7.74 & \textless .00001 \\
    29 & lid, soap\_dispenser & 33 & 100.00 & 80.47 & 4.38 & 2.14 & 4.15 & 1.74 & -5.92 & \textless .00001 \\
    30 & teapot, saucepan & 27 & 85.19 & 49.93 & 2.52 & 1.05 & 2.23 & 0.00 & -4.28 & \textless .00001 \\
    \hline
    36 & tap, crapper & 23 & 91.30 & 70.78 & 3.24 & 1.75 & 2.82 & 1.29 & -3.59 & \textless .00001 \\
    41 & open\_fireplace, coffee\_table & 31 & 80.65 & 15.11 & 2.03 & 0.14 & 2.12 & 0.00 & -7.15 & \textless .00001 \\
    43 & central\_reservation & 40 & 97.50 & 85.42 & 7.43 & 3.71 & 8.08 & 3.60 & -5.94 & \textless .00001 \\
    48 & road & 42 & 100.00 & 74.46 & 6.15 & 2.68 & 6.65 & 2.30 & -7.78 & \textless .00001 \\
    49 & footboard, chain & 32 & 84.38 & 66.41 & 2.63 & 1.67 & 2.30 & 1.17 & -2.58 & 0.0049 \\
    \hline
    51 & road, car & 21 & 100.00 & 47.65 & 5.32 & 1.52 & 5.62 & 0.00 & -6.03 & \textless .00001 \\
    54 & skyscraper & 39 & 100.00 & 71.78 & 4.14 & 1.61 & 4.08 & 1.12 & -7.60 & \textless .00001 \\
    56 & flusher, soap\_dish & 53 & 92.45 & 64.29 & 3.47 & 1.48 & 3.08 & 0.86 & -6.47 & \textless .00001 \\
    57 & shower\_stall, screen\_door & 34 & 97.06 & 32.31 & 2.60 & 0.61 & 2.53 & 0.00 & -7.55 & \textless .00001 \\
    63 & edifice, skyscraper & 45 & 88.89 & 48.38 & 2.41 & 0.83 & 2.36 & 0.00 & -6.73 & \textless .00001 \\
    \hline
\end{tabular}
\end{footnotesize}
\caption{Evaluation details as discussed in Section~\ref{sec:eval}.
Images: Number of images used for evaluation.
\# Activations: (targ(et)): Percentage of target images activating the neuron (i.e.,
activation at least 80\% of this neuron's activation maximum);
(\mbox{non-t}):
Same for all
other images used in the evaluation. Mean/Median (targ(et)/non-t(arget)): Mean/median activation value for
target and non-target images, respectively.}
\label{tab:evaluation}
\end{table*}

We consider each neuron-label pair in each row in Table~\ref{tab:evaluation}, e.g., for neuron 1, the hypothesis is that this neuron activates more strongly for images retrieved using the keyword ``cross\_walk'' than for images retrieved using other keywords. The corresponding null hypothesis is that activation values are \emph{not} different.
Table~\ref{tab:evaluation} shows the 20 hypotheses to test, corresponding to the 20 neurons with confirmed labels
-- recall that a double label such as neuron 16's ``mountain, bushes'' is treated as one label consisting of the conjunction of the two keywords. 

There is no reason to assume that activation values would follow a normal distribution, or that the preconditions of the central limit theorem would be satisfied. We therefore base our statistical assessment on the Mann-Whitney U test~\cite{McKnight2010MannWhitneyUT} which is a non-parametric test that does not require a normal distribution. Essentially, by comparing the ranks of the observations in the two groups, the test allows us to determine if there is a statistically significant difference in the activation percentages between the target and non-target labels.

The resulting z-scores and p-values are shown in Table~\ref{tab:evaluation}. Of the 20 null hypotheses,
19 are rejected at $p < 0.05$, but most (all except neurons 0, 18 and 49) are rejected at much
	lower p-values. Only neuron 0's null hypothesis could not be rejected. The Non-Target \%
	column of Table~\ref{tab:listofConcepts-abbrv}
	provides some insight into the results for neurons 0, 18 and 49: target and non-target values
	for these neurons are closer to each other -- the difference is particularly small for neuron 0. Likewise, differences between target and non-target values for mean activation values and median activation values in Table~\ref{tab:evaluation} are smaller for these neurons. This hints at ways to improve label hypothesis generation or confirmation, and we will discuss this and other ideas for further improvement below under possible future work.

\begin{figure*}[tb]
\centering
\includegraphics[clip, trim=0.5in 6.3in 0.5in 0.5in, width=0.99\textwidth]{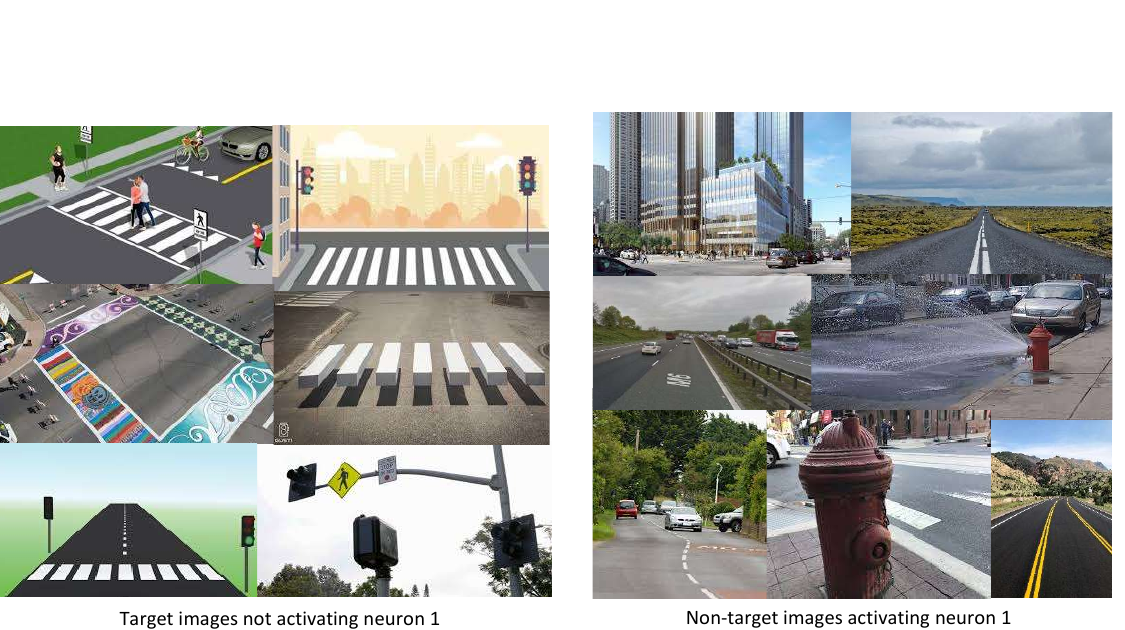} 
\caption{Examples of some Google images used: target images (``cross\_walk'') that did not
activate the neuron; non-target images from labels like ``central\_reservation,'' ``road and
car,'' and ``fire\_hydrant'' that activated the neuron.}
\label{fig2}
\end{figure*}

For our running example (neuron 1), we use the remaining 47 target images (20\% of the 165 images retrieved during the label hypothesis confirmation step) for the statistical analysis. 43 of these images (91.49\%) activate the neuron
at $\ge$ 8.72 (80\% of its maximum activation value of 10.90), with a mean and median activation of 4.17 and 4.13, respectively. Of all other images (non-target images) used in the evaluation (the sum of the numbers in the image column in Table~\ref{tab:evaluation} minus 47), only 28.94\% activate neuron 1 at $\ge$ 8.72, for a mean of 0.67 and a median of 0.00. The Mann-Whitney U test yields a z-score of -8.92 and $p<0.00001$, thus rejecting the null hypothesis that activation values for target and non-target images are \emph{not} different. In addition, the negative z-score indicates that the activation values for non-target images are indeed lower than for the target images. Figure~\ref{fig2} shows examples of target images that do not activate neuron 1 (left) and non-target images that do activate it (right).

The Mann-Whitney U results show that, for most neurons listed in Table~\ref{tab:evaluation} (with $p<0.00001$), activation values for target images are \emph{overwhelmingly} higher than for non-target images. The negative z-scores with high absolute values informally indicate the same, as do the mean and median values.
Neurons 16 and 49, for which the hypotheses also hold but with $p<0.05$ and $p<0.01$, respectively, still exhibit statistically significant higher activation values for target than for non-target images, but not overwhelmingly so. This can also be informally seen from lower absolute values of the z-scores, and from smaller differences between the means and the medians.

\section{Discussion}
\label{sec:discussion}

While the statistical analysis clearly supports the viability of our approach as carried out,
we can see from Table~\ref{tab:evaluation} that there is still significant activation of neurons by non-target images, leaving room for refinement. Ideally, we would be able to arrive at confirmed neuron labels where the number of non-target activations (Table~\ref{tab:evaluation}, \# Activations (\%) non-t -- column 5)
is very low while the number of target activations (\# Activations(\%) targ -- column 4)
remains high.
For example, neuron 16 is always activated by target images and is only activated by 25\% of non-target images, meaning that we can use the neuron activation to predict (with relatively high certainty) whether or not mountains and bushes are in the image. In contrast, for neuron 29 -- with a much higher non-target value -- we can be much less certain if an image activating the neuron indeed contains lid and soap dispenser. High certainty, i.e., high target and low non-target values, would provide highly accurate explanations of system behavior. At the same time, however, the data collected during the label generation step, in particular that in the Target \% and Non-Target \% columns of Table~\ref{tab:listofConcepts-abbrv}, can already be used as a proxy for the certainty we can attach to a detection.

It is instructive to have another look at our example neuron 1. The images depicted on the
left in Figure~\ref{fig2} -- target images not activating the neuron -- are mostly
computer-generated as opposed to photographic images as in the ADE20K dataset. The lower right
image does not actually show the ground at the crosswalk, but mostly sky and only indirect evidence for a crosswalk by
means of signage, which may be part of the reason why the neuron does not activate. 
The right-hand images are non-target images that activate the neuron. We
may conjecture that other road elements, prevalent in these pictures, may have triggered the
neuron. We also note that several of these images show bushes or plants, which is
particularly interesting because the ECII response with the third-highest coverage score is
``bushes, bush'' with a coverage score of 0.993 and 48.052\% of images retrieved using this
label actually activate the neuron (the second response for this neuron is also ``cross\_walk'').

Our results point to promising 
future research directions, including studying ensemble activation, analyzing different hidden layers,
transfer to other application scenarios, and application to other deep learning architectures. Other possible avenues involve additional strengthening of our results including detailed analyses and exploration -- and thus optimization -- of parameters that were often chosen ad-hoc for this paper. We discuss some of them below, in sequence of appearance in the paper. Additional results along these lines can be found in Appendix~\ref{appendix}.

The choice of background knowledge -- based on objects appearing in the images -- was mostly one of convenience, as suitable (large-scale) datasets were available that would serve the purpose of this study. However it is conceivable -- if not likely -- that neuron activations are caused not (only) by the positioning of (types of) objects, but also by other image features such as prevalence, (relative) positioning of lines or round shapes, contrasts across the image, colors,
etc., some of which may be numerical in nature. It is of course possible to compile corresponding background knowledge for Concept Induction analysis, together with appropriate annotations of images, and we assume that results can be strengthened by making use of a suitably designed knowledge base. It should also be noted that the background knowledge (and mappings)
are not
tightly curated, but -- because of scale -- their generation was based on heuristics, and thus contains some imperfections. More tightly quality controlled background knowledge should further improve our results. Background knowledge bases that make more sophisticated use of description logic axiomatization (together with the DL-Learner Concept Induction system~\cite{DBLP:journals/ml/LehmannH10} or new heuristics that would need developing if at large scale) should also strengthen the results.

Regarding label hypothesis generation (Section~\ref{subsec:label-hypotheses}), our use of 80\% and 20\% of the maximum activation value for each neuron as cut-offs for selecting the images that go into the Concept Induction analysis can likely be refined, they were mostly selected ad-hoc. Use of coverage score to select the top-ranking Concept Induction system response could be replaced by others such as f-measure. We have also, so far, ignored lower ranked responses by the Concept Induction system, although often their coverage scores are very close to that of the top ranked response. Exploring ways to leverage top-n ranked responses should lead to ways to improve (i.e., increase) the target vs. non-target activation gap.
Further refinement may be possible by also taking the values in the Non-Target \% column
into consideration, or incorporating statistical analysis
at this stage as well.

\section{Conclusion and Future Work}
\label{conclusion}

We have demonstrated that our approach using Concept Induction and large-scale background knowledge leads to meaningful labeling of hidden neuron activations, as confirmed by our statistical analysis. To the best of our knowledge, this approach is new, and in particular the use of large-scale background knowledge for this purpose -- which means that label categories are not restricted to a few pre-selected terms -- has not been explored before.

A major direction for future work is analysing activations of neuron ensembles rather than single neurons -- intuitively, information would often be distributed over several simultaneously activated neurons. Scale is a major obstacle to this type of investigation, as even with only, say, 64 hidden neurons in a layer, there are already about $2^{64}$ possible neuron ensembles that could be investigated, i.e., brute-force analysis methods are not feasible in most contexts, and better ways to navigate this search space will have to be found. Possible refinements are to
combine the
neurons that activate for semantically related labels (e.g., neurons 0, 22, 26, 54, and 63 in Table~\ref{tab:listofConcepts}) and/or 
take the top-n ranked responses from the Concept Induction system into account.

Eventually, our line of work aims at comprehensive and conclusive hidden layer analysis for deep learning systems, so that, after analysis, it is possible to ``read off'' from the activations, (some of) the implicit features of the input that the network has detected, thus opening up avenues to really explaining the system's input-output behavior.

\subsubsection*{Acknowledgments}

This research has been supported by the National Science Foundation under Grant No. 2033521.

\newpage
\bibliography{refs}

\newpage
\input{appendix}

\end{document}

%% file: appendix.tex

\appendix
\section{Technical Appendix}
\label{appendix}

Table~\ref{tab:listofConcepts} is the full version of Table~\ref{tab:listofConcepts-abbrv}, which, due to page restrictions, we could not fit into the main paper. 

In the rest of the appendix, we will first present results concerning second and third ranked ECII responses (Section~\ref{subsec:alternativetop-n}). Then (in Section~\ref{subsec:earlier-exploratory}) we include some earlier exploratory data (on a preliminary system), that informed our study.

\begin{table*}[t!]
\centering
\begin{footnotesize}
\begin{tabular}{clrrrr}
    Neuron \# & Obtained Label(s) & Images & Coverage & Target \% & Non-Target \% \\
    \hline
    \textbf{0} & \textbf{building} & \textbf{164} & \textbf{0.997} & \textbf{89.024} & \textbf{72.328} \\
    \textbf{1} & \textbf{cross\_walk} & \textbf{186} & \textbf{0.994} & \textbf{88.710} & \textbf{28.923} \\
    \textbf{3} & \textbf{night\_table} & \textbf{157} & \textbf{0.987} & \textbf{90.446} & \textbf{56.714} \\
    6 & dishcloth, toaster & 106 & 0.999 & 16.038 & 39.078 \\
    7 & toothbrush, pipage & 112 & 0.991 & 75.893 & 59.436 \\
    \hline
    \textbf{8} & \textbf{shower\_stall, cistern} & \textbf{136} & \textbf{0.995} & \textbf{100.000} & \textbf{53.186} \\
    11 & river\_water & 157 & 0.995 & 31.847 & 22.309 \\
    12 & baseboard, dish\_rag & 108 & 0.993 & 75.926 & 48.248 \\
    14 & rocking\_horse, rocker & 86 & 0.985 & 54.651 & 47.816 \\
    \textbf{16} & \textbf{mountain, bushes} & \textbf{108} & \textbf{0.995} & \textbf{87.037} & \textbf{24.969} \\
    \hline
    17 & stem & 133 & 0.993 & 30.827 & 31.800 \\
    \textbf{18} & \textbf{slope} & \textbf{139} & \textbf{0.983} & \textbf{92.086} & \textbf{69.919} \\
    \textbf{19} & \textbf{wardrobe, air\_conditioning} & \textbf{110} & \textbf{0.999} & \textbf{89.091} & \textbf{65.034} \\
    20 & fire\_hydrant & 158 & 0.990 & 5.696 & 13.233 \\
    \textbf{22} & \textbf{skyscraper} & \textbf{156} & \textbf{0.992} & \textbf{99.359} & \textbf{54.893} \\
    \hline
    23 & fire\_escape & 162 & 0.996 & 61.111 & 18.311 \\
    25 & spatula, nuts & 126 & 0.999 & 2.381 & 0.883 \\
    26 & skyscraper, river & 112 & 0.995 & 77.679 & 35.489 \\
    27 & manhole, left\_arm & 85 & 0.996 & 35.294 & 26.640 \\
    28 & flooring, fluorescent\_tube & 115 & 1.000 & 38.261 & 33.198 \\
    \hline
    \textbf{29} & \textbf{lid, soap\_dispenser} & \textbf{131} & \textbf{0.998} & \textbf{99.237} & \textbf{78.571} \\
    \textbf{30} & \textbf{teapot, saucepan} & \textbf{108} & \textbf{0.998} & \textbf{81.481} & \textbf{47.984} \\
    \textbf{31} & fire\_escape & 162 & 0.961 & 77.160 & 63.147 \\
    33 & tanklid, slipper & 81 & 0.987 & 41.975 & 30.214 \\
    34 & left\_foot, mouth & 110 & 0.994 & 20.909 & 49.216 \\
    \hline
    35 & utensils\_canister, body & 111 & 0.999 & 7.207 & 11.223 \\
    \textbf{36} & \textbf{tap, crapper} & \textbf{92} & \textbf{0.997} & \textbf{89.130} & \textbf{70.606} \\
    37 & cistern, doorcase & 101 & 0.999 & 21.782 & 24.147 \\
    38 & letter\_box, go\_cart & 125 & 0.999 & 28.000 & 31.314 \\
    39 & side\_rail & 148 & 0.980 & 35.811 & 34.687 \\
    \hline
    40 & sculpture, side\_rail & 119 & 0.995 & 25.210 & 21.224 \\
    \textbf{41} & \textbf{open\_fireplace, coffee\_table} & \textbf{122} & \textbf{0.992} & \textbf{88.525} & \textbf{16.381} \\
    42 & pillar, stretcher & 117 & 0.998 & 52.137 & 42.169 \\
    \textbf{43} & \textbf{central\_reservation} & \textbf{157} & \textbf{0.986} & \textbf{95.541} & \textbf{84.973} \\
    44 & saucepan, dishrack & 120 & 0.997 & 69.167 & 36.157 \\
    \hline
    46 & Casserole & 157 & 0.999 & 45.223 & 36.394 \\
    \textbf{48} & \textbf{road} & \textbf{167} & \textbf{0.984} & \textbf{100.000} & \textbf{73.932} \\
    \textbf{49} & \textbf{footboard, chain} & \textbf{126} & \textbf{0.982} & \textbf{88.889} & \textbf{66.702} \\
    50 & night\_table & 157 & 0.972 & 65.605 & 62.735 \\
    \textbf{51} & \textbf{road, car} & \textbf{84} & \textbf{0.999} & \textbf{98.810} & \textbf{48.571} \\
    \hline
    53 & pylon, posters & 104 & 0.985 & 11.538 & 17.332 \\
    \textbf{54} & \textbf{skyscraper} & \textbf{156} & \textbf{0.987} & \textbf{98.718} & \textbf{70.432} \\
    \textbf{56} & \textbf{flusher, soap\_dish} & \textbf{212} & \textbf{0.997} & \textbf{90.094} & \textbf{63.552} \\
    \textbf{57} & \textbf{shower\_stall, screen\_door} & \textbf{133} & \textbf{0.999} & \textbf{98.496} & \textbf{31.747} \\
    58 & plank, casserole & 80 & 0.998 & 3.750 & 3.925 \\
    \hline
    59 & manhole, left\_arm & 85 & 0.994 & 35.294 & 21.589 \\
    60 & paper\_towels, jar & 87 & 0.999 & 0.000 & 1.246 \\
    61 & ornament, saucepan & 102 & 0.995 & 43.137 & 17.274 \\
    62 & sideboard & 100 & 0.991 & 21.000 & 29.734 \\
    \textbf{63} & \textbf{edifice, skyscraper} & \textbf{178} & \textbf{0.999} & \textbf{92.135} & \textbf{48.761}\\
 
    \hline
\end{tabular}
\end{footnotesize}
\caption{Unabridged version of Table~\ref{tab:listofConcepts-abbrv} The omitted neurons were not activated
by any image, i.e., their maximum activation value was 0. Images: Number of images used per label. Target \%: Percentage of target images activating the neuron above 80\% of its maximum activation. Non-Target \%: The same, but for all other images.
\textbf{Bold} denotes the 20 neurons whose labels are considered confirmed.}
\label{tab:listofConcepts}
\end{table*}

\subsection{Second- and third-ranked ECII responses}
\label{subsec:alternativetop-n}

The Concept Induction system ECII does not return only one result, but instead a list of results ranked (in our case) by coverage. In the main body of the paper we concerned ourselves only with the first ECII response, but we observed that neurons sometimes also activate for non-target image inputs, i.e., the label derived from the first ECII response gives only a partial picture. We will now look at second and third responses, essentially repeating the analysis done for the first responses in the main paper.

\subsubsection{Second ECII Responses}
\label{subsubsec:top-2}

Following the approach outlined in the main text, we sorted the ECII responses by their coverage scores. For each neuron, we now considered the label hypothesis corresponding to the second-highest response obtained from running ECII on both the positive and negative sets of images, as described in Section~\ref{subsec:label-hypotheses}, using the background knowledge described in Section~\ref{subsec:CI}. To confirm or reject these hypotheses we follow the criterion described in Section~\ref{subsec:label-confirmation}, i.e., we retrieve images from Google Images with the target labels as search keywords. Table~\ref{tab:top2_listofConcepts} lists the results for each neuron, including the label hypothesis associated with the second-highest ECII response, along with the coverage score of the corresponding ECII response and the number of images used in the hypothesis confirmation step for each label. Following the main text, the ``Target \%'' column in Table~\ref{tab:top2_listofConcepts} displays the percentage of target images that activated each neuron to at least 80\% of its maximum activation value. A target label for a neuron is defined as confirmed if it activates (with at least 80\% of its maximum activation value) for at least 80\% of its target images, regardless of how much or how often it activates for non-target images. Confirmed labels are highlighted in boldface in Table~\ref{tab:top2_listofConcepts}.

\begin{table*}[t!]
\centering
\begin{tabular}{clcccc}

    Neuron \# & ECII Concept(s) & Images & Coverage & Target \% & Non-Target \% \\
    \hline
    \textbf{0} & \textbf{building, dome} & \textbf{250} & \textbf{0.997} & \textbf{90.4} & \textbf{78.185} \\
    \textbf{1} & \textbf{cross\_walk} & \textbf{187} & \textbf{0.994} & \textbf{88.770} & \textbf{28.241} \\
    \textbf{3} & \textbf{pillow} & \textbf{168} & \textbf{0.984} & \textbf{98.214} & \textbf{61.250} \\
    6 & dishcloth, paper\_towel & 221 & 0.999 & 18.100 & 36.985 \\
    \textbf{7} & \textbf{clamp\_lamp} & \textbf{144} & \textbf{0.991} & \textbf{95.139} & \textbf{59.504} \\
    \hline
    8 & dishcloth, saucepan & 242 & 0.995 & 57.851 & 52.196 \\
    11 & river & 159 & 0.995 & 23.270 & 22.336 \\
    12 & dishrag & 112 & 0.993 & 76.786 & 49.488 \\
    14 & display\_board, marker & 341 & 0.985 & 30.499 & 54.797 \\
    16 & bushes, bush & 154 & 0.995 & 49.351 & 20.690 \\
    \hline
    17 & opening, knife & 281 & 0.993 & 12.456 & 30.611 \\
    \textbf{18} & \textbf{slope} & \textbf{140} & \textbf{0.983} & \textbf{92.143} & \textbf{64.503} \\
    19 & wardrobe, closet & 263 & 0.999 & 79.468 & 67.042 \\
    20 & roof & 90 & 0.990 & 27.778 & 9.221 \\
    \textbf{22} & \textbf{skyscraper} & \textbf{157} & \textbf{0.992} & \textbf{99.363} & \textbf{62.185} \\
    \hline
    23 & fire\_escape & 163 & 0.996 & 61.350 & 21.363 \\
    25 & whisk, solid\_food & 200 & 0.999 & 0 & 1.526 \\
    26 & river\_water & 158 & 0.995 & 26.582 & 41.810 \\
    27 & manhole, head & 302 & 0.996 & 35.099 & 28.372 \\
    28 & chair, podium & 318 & 1.000 & 21.069 & 26.297 \\
    \hline
    \textbf{29} & \textbf{faucet, flusher} & \textbf{302} & \textbf{0.998} & \textbf{95.695} & \textbf{78.562} \\
    30 & posting, saucepan & 270 & 0.998 & 68.889 & 44.696 \\
    31 & wires & 110 & 0.961 & 72.727 & 70.369 \\
    33 & toilet\_brush, slipper & 325 & 0.987 & 32.923 & 37.090 \\
    34 & left\_hand, crane & 176 & 0.994 & 56.818 & 54.899 \\
    \hline
    35 & range, vent & 210 & 0.999 & 39.048 & 10.604 \\
    \textbf{36} & \textbf{tap, shower\_screen} & \textbf{320} & \textbf{0.997} & \textbf{86.25} & \textbf{72.584} \\
    37 & cistern, lid & 326 & 0.999 & 21.166 & 23.277 \\
    38 & baby\_buggy, baby\_carriage & 260 & 0.999 & 32.308 & 29.024 \\
    39 & side\_rail & 148 & 0.980 & 35.811 & 34.687 \\
    \hline
    40 & sculpture, footboard & 325 & 0.995 & 40.308 & 18.563 \\
    41 & open\_fireplace, seat\_base & 335 & 0.992 & 43.284 & 18.297 \\
    42 & pillar, arm\_support & 333 & 0.998 & 56.156 & 45.177 \\
    \textbf{43} & \textbf{central\_reservation} & \textbf{158} & \textbf{0.986} & \textbf{95.570} & \textbf{90.141} \\
    44 & dishcloth, saucepan & 242 & 0.997 & 57.438 & 32.037 \\
    \hline
    46 & ornament, saucepan & 272 & 0.999 & 36.029 & 33.696 \\
    \textbf{48} & \textbf{route} & \textbf{172} & \textbf{0.984} & \textbf{100} & \textbf{80.834} \\
    49 & desk\_lamp, candle & 319 & 0.981 & 75.235 & 70.135 \\
    \textbf{50} & \textbf{pillow} & \textbf{168} & \textbf{0.966} & \textbf{99.405} & \textbf{66.834} \\
    \textbf{51} & \textbf{route, car} & \textbf{312} & \textbf{0.999} & \textbf{92.628} & \textbf{47.408} \\
    \hline
    53 & bushes, bush & 154 & 0.985 & 12.338 & 13.749 \\
    \textbf{54} & \textbf{skyscraper} & \textbf{157} & \textbf{0.987} & \textbf{98.726} & \textbf{74.687} \\
    56 & bannister, hand\_rail & 312 & 0.997 & 36.538 & 63.751 \\
    57 & cistern, screen\_door & 317 & 0.999 & 45.741 & 30.182 \\
    58 & baseboard, board & 244 & 0.998 & 3.689 & 3.695 \\
    \hline
    59 & manhole & 170 & 0.994 & 50 & 25.017 \\
    60 & dishcloth, pail & 242 & 0.999 & 3.719 & 2.463 \\
    61 & ornamentation, casserole & 317 & 0.995 & 32.492 & 17.445 \\
    62 & sideboard & 101 & 0.991 & 21.782 & 36.283 \\
    \textbf{63} & \textbf{building, skyscraper} & \textbf{321} & \textbf{0.999} & \textbf{94.393} & \textbf{51.612}\\
    \hline
\end{tabular}
\caption{Activation percentage for label hypothesis for each neuron, derived from second ECII responses. The omitted neurons were not activated by any image, i.e., their maximum activation value was 0. Images: Number of images used per label. Target \%: Percentage of target images activating the neuron above 80\% of its maximum activation. Non-Target: The same, but for all other images.
\textbf{Bold} denotes the 14 neurons whose labels are considered confirmed.}
\label{tab:top2_listofConcepts}
\end{table*}

\begin{table*}[t]
\centering
\begin{tabular}{clc|cc|cc|cc|cr}
     \multicolumn{1}{l}{Neuron \#} & \multicolumn{1}{l}{Label(s)} & \multicolumn{1}{c}{Images} & \multicolumn{2}{c}{\# Activations (\%)} & \multicolumn{2}{c}{Mean} & \multicolumn{2}{c}{Median} & \multicolumn{1}{c}{z-score} & \multicolumn{1}{c}{p-value}\\
     \hline
     &  & & targ & non-t & targ & non-t & targ & non-t &  & \\
    \hline
    0 & building, dome & 63 & 90.48 & 78.05 & 2.55 & 1.60 & 2.37 & 1.832 & -2.83 & .002327 \\
    1 & cross\_walk & 47 & 91.49 & 27.75 & 4.13 & 0.00 & 4.17 & 0.452 & -11.94 & \textless   .00001 \\
    3 & pillow & 42 & 92.86 & 61.46 & 3.89 & 0.62 & 3.62 & 1.171 & -7.74 & \textless   .00001 \\
    7 & clamp\_lamp & 36 & 97.22 & 59.18 & 3.10 & 0.54 & 2.76 & 1.326 & -5.71 & \textless   .00001 \\
    18 & slope & 35 & 91.43 & 64.19 & 1.44 & 0.73 & 1.59 & 1.195 & -2.89 & .001957 \\
    22 & skyscraper & 39 & 97.44 & 62.38 & 4.42 & 0.79 & 3.97 & 1.419 & -7.78 & \textless   .00001 \\
    29 & faucet, flusher & 76 & 97.37 & 78.41 & 4.13 & 1.74 & 4.25 & 2.174 & -7.86 & \textless   .00001 \\
    \hline
    36 & tap, shower\_screen & 80 & 86.25 & 71.43 & 3.18 & 1.08 & 3.19 & 1.574 & -6.64 & \textless   .00001 \\
    43 & central\_reservation & 40 & 97.50 & 89.92 & 8.08 & 3.70 & 7.43 & 3.699 & -6.05 & \textless   .00001 \\
    48 & route & 43 & 100.00 & 80.66 & 4.23 & 2.23 & 4.42 & 2.531 & -5.99 & \textless   .00001 \\
    50 & pillow & 42 & 97.62 & 67.76 & 4.47 & 0.91 & 4.43 & 1.415 & -8.63 & \textless   .00001 \\
    51 & route, car & 78 & 97.44 & 46.55 & 2.91 & 0.00 & 3.29 & 1.122 & -10.67 & \textless   .00001 \\
    54 & skyscraper & 39 & 100.00 & 75.24 & 4.08 & 1.34 & 4.14 & 1.730 & -7.43 & \textless   .00001 \\
    63 & building, skyscraper & 81 & 92.59 & 51.45 & 2.61 & 0.08 & 2.75 & 0.752 & -10.89 & \textless   .00001 \\
    \hline
\end{tabular}
\caption{Evaluation details as discussed in Section~\ref{subsubsec:top-2} for second ECII responses.
Images: number of images used for evaluation.
\# Activations: (targ(et)): Percentage of target images activated the neuron (i.e., activation at least 80\% of this neuron's activation maximum); (non-t(arget)): Same for all other images used in the evaluation. Mean/Median (targ(et)/non-t(arget)): Mean/median activation value for target and non-target images, respectively.}
\label{tab:top2_evaluation}
\end{table*}

Table~\ref{tab:top2_evaluation} presents the statistical evaluation of the 14 confirmed label hypotheses, which were established through the analysis of the second ECII responses for each neuron in Table~\ref{tab:top2_listofConcepts}. The statistical evaluation follows the same approach as in Section~\ref{sec:eval}, using Mann-Whitney U test z-scores and p-values to assess the significance of the differences between the target and non-target activations for each hypothesis. Table~\ref{tab:evaluation} shows various metrics including target and non-target activations, target and non-target medians, target and non-target means, z-scores, and p-values. 

The results in Table~\ref{tab:top2_evaluation} show that 12 out of 14 hypotheses are statistically significant with $p < 0.00001$, but all hypotheses are statistically significant at $p < 0.05$. The z-scores for the 14 hypotheses are all negative, indicating that the target activations are higher than the non-target activations. Overall, the results in Table~\ref{tab:top2_evaluation} provide strong evidence for the 14 hypotheses. 

The results are strongly suggestive for follow-on investigations. We note that the coverage scores listed in Table~\ref{tab:top2_listofConcepts} are still very high, and 14 of the label hypotheses are both confirmed and statistically validated, a smaller but still significant number. Comparing labels in Table~\ref{tab:top2_evaluation} with those in Table~\ref{tab:evaluation}, we note that some are duplicated (e.g., ``skyscraper'' for neuron 22), while others appear related (like ``pillow'' and ``night\_table'' for neuron 3). Notably, for some neurons we could both confirm and statistically validate the label hypotheses although the label from the first ECII response could not be confirmed (e.g., ``pillow'' for neuron 50 was confirmed, while the first ECII response ``night\_table'' listed in Table~\ref{tab:evaluation} was not confirmed). We hypothesize that combining first, second, and third (see below) responses will provide labels that have higher target activations but lower non-target activations, compared to taking only the first ECII response. The exact way of selecting top scoring responses, finding optimal activation cut-offs for confirming labels, and combining labels, will require follow-up investigation.

\subsubsection{Third ECII Responses}
\label{subsubsec:top-3}
For the third ECII responses, we follow the exact same workflow and analysis as for the first and second ECII responses.  The results are shown in Tables~\ref{tab:top3_listofConcepts} and~\ref{tab:top3_evaluation}. Table~\ref{tab:top3_listofConcepts} lists the label hypotheses of each neuron along with the count of target images used, coverage scores, and target and non-target activations. Table~\ref{tab:top3_evaluation} presents the results of the statistical evaluation of the 13 confirmed label hypotheses
 including target and non-target activations, means, medians, z-scores, and p-values for each neuron. All 13 passed the statistical validation at $p < 0.001$.

As for the second ECII responses, we note that coverage scores remain high, and observations made for the second ECII responses above hold accordingly. 

For example, we now obtain a confirmed and statistically validated label hypothesis for neuron 62, ``buffet,'' (Table~\ref{tab:top3_listofConcepts}) while the second and first ECII responses, both ``sideboard,'' (Tables~\ref{tab:top2_listofConcepts} and~\ref{tab:listofConcepts}) were only activated by 21.782\% respectively 21.000\% of target images (and as such not confirmed). As already mentioned, ways of meaningfully combining different ECII responses will be subject to future investigations.

\begin{table*}[t!]
\centering
\begin{tabular}{clcccc}
    Neuron \# & Obtained Label(s) & Images & Coverage & Target \% & Non-Target \% \\
    \hline
    0 & sky, boat & 201 & 0.997 & 76.617 & 72.256 \\
    1 & bushes, bush & 154 & 0.993 & 48.052 & 27.408 \\
    \textbf{3} & \textbf{pillow} & \textbf{168} & \textbf{0.984} & \textbf{98.214} & \textbf{63.703} \\
    6 & baseboard, paper\_towels & 249 & 0.999 & 53.414 & 31.705 \\
    \textbf{7} & \textbf{clamp\_lamp, clamp} & \textbf{307} & \textbf{0.991} & \textbf{95.765} & \textbf{63.456} \\
    \hline
    8 & towel\_horse, cistern & 258 & 0.995 & 64.341 & 56.170 \\
    11 & dock, river\_water & 261 & 0.994 & 34.100 & 15.146 \\
    12 & dishcloth & 125 & 0.993 & 69.600 & 53.742 \\
    14 & marker & 185 & 0.985 & 21.622 & 60.449 \\
    \textbf{16} & \textbf{mountain, bush} & \textbf{108} & \textbf{0.995} & \textbf{87.037} & \textbf{24.969} \\
    \hline
    17 & napkin\_ring & 115 & 0.993 & 43.478 & 30.194 \\
    \textbf{18} & \textbf{field} & \textbf{159} & \textbf{0.983} & \textbf{91.824} & \textbf{65.333} \\
    \textbf{19} & \textbf{closet, air\_conditioning} & \textbf{267} & \textbf{0.999} & \textbf{86.891} & \textbf{71.054} \\
    20 & fire\_hydrant & 158 & 0.990 & 5.696 & 9.771 \\
    22 & left\_hand, right\_hand & 169 & 0.987 & 68.639 & 62.675 \\
    \hline
    23 & lamps, bulletin\_board & 250 & 0.995 & 16.400 & 16.945 \\
    25 & whisk, pot\_rack & 225 & 0.999 & 8.889 & 1.021 \\
    26 & river & 159 & 0.995 & 13.208 & 40.658 \\
    27 & head, right\_arm & 234 & 0.995 & 22.650 & 27.417 \\
    28 & chair, ceiling & 353 & 1.000 & 28.895 & 32.904 \\
    \hline
    \textbf{29} & \textbf{potty, flusher} & \textbf{305} & \textbf{0.998} & \textbf{88.525} & \textbf{76.830} \\
    30 & cup, dishcloth & 211 & 0.998 & 36.493 & 44.479 \\
    31 & wires & 110 & 0.961 & 72.727 & 66.657 \\
    33 & toilet\_brush & 141 & 0.987 & 24.823 & 36.881 \\
    34 & crane, right\_foot & 190 & 0.994 & 47.895 & 53.201 \\
    \hline
    35 & skirting\_board, vent & 269 & 0.999 & 19.331 & 8.686 \\
    36 & crapper, throne & 257 & \textbf{0.997} & 58.366 & \textbf{79.198} \\
    37 & tap, shower\_screen & 320 & 0.999 & 45.938 & 23.486 \\
    38 & pram, mailbox & 317 & 0.999 & 23.344 & 29.155 \\
    39 & left\_hand, crane & 176 & 0.979 & 24.432 & 38.702 \\
    \hline
    40 & bucket, phone & 291 & 0.994 & 18.213 & 19.252 \\
    41 & seat\_base, coffe\_table & 325 & 0.989 & 34.154 & 16.018 \\
    42 & h-stretcher, bookcase & 297 & 0.998 & 39.394 & 45.984 \\
    \textbf{43} & \textbf{mountain} & \textbf{158} & \textbf{0.982} & \textbf{99.367} & \textbf{88.516} \\
    44 & cup, dishrack & 180 & 0.997 & 75.556 & 37.029 \\
    \hline
    46 & ornamentation, saucepan & 273 & 0.999 & 40.659 & 29.727 \\
    \textbf{48} & \textbf{road} & \textbf{168} & \textbf{0.984} & \textbf{100} & \textbf{76.789} \\
    49 & candle, chain & 286 & 0.981 & 78.671 & 72.968 \\
    \textbf{50} & \textbf{pillow} & \textbf{168} & \textbf{0.966} & \textbf{99.405} & \textbf{72.085} \\
    \textbf{51} & \textbf{road, automobile} & \textbf{336} & \textbf{0.998} & \textbf{92.560} & \textbf{41.466} \\
    \hline
    53 & hill, tank & 262 & 0.985 & 36.641 & 15.186 \\
    \textbf{54} & \textbf{hedgerow, hedge} & \textbf{249} & \textbf{0.985} & \textbf{91.165} & \textbf{68.527} \\
    56 & handrail, shoer\_stall & 301 & 0.997 & 59.801 & 66.330 \\
    57 & toilet\_brush, cistern & 300 & 0.998 & 34.333 & 33.016 \\
    58 & mopboard, plank & 161 & 0.998 & 5.590 & 2.164 \\
    \hline
    59 & manhole & 170 & 0.994 & 50.000 & 25.017 \\
    60 & baseboard, soap\_bottle & 289 & 0.999 & 0.692 & 1.056 \\
    61 & fryingpan, sash & 263 & 0.995 & 22.814 & 17.121 \\
    \textbf{62} & \textbf{buffet} & \textbf{61} & \textbf{0.991} & \textbf{83.607} & \textbf{32.714} \\
    \textbf{63} & \textbf{skyscraper} & \textbf{157} & \textbf{0.998} & \textbf{99.363} & \textbf{52.947}\\
    \hline
\end{tabular}
\caption{Activation percentage for label hypothesis for each neuron, derived from third ECII responses. The omitted neurons were not activated by any image, i.e., their maximum activation value was 0.
Images: Number of images used per label. Target \%: Percentage of target images activating the neuron above 80\% of its maximum activation. Non-Target \%: The same, but for all other images.
\textbf{Bold} denotes the 13 neurons whose labels are considered confirmed.}
\label{tab:top3_listofConcepts}
\end{table*}

\begin{table*}[t]
\centering
\begin{tabular}{clc|cc|cc|cc|cr}
     \multicolumn{1}{l}{Neuron \#} & \multicolumn{1}{l}{Label(s)} & \multicolumn{1}{c}{Images} & \multicolumn{2}{c}{\# Activations (\%)} & \multicolumn{2}{c}{Mean} & \multicolumn{2}{c}{Median} & \multicolumn{1}{c}{z-score} & \multicolumn{1}{c}{p-value}\\
     \hline
     &  & & targ & non-t & targ & non-t & targ & non-t &  & \\
    \hline
    3 & pillow & 42 & 92.86 & 64.05 & 3.62 & 0.80 & 3.89 & 0.80 & -7.29 & \textless .00001 \\
    7 & clamp\_lamp, clamp & 77 & 96.10 & 65.04 & 2.73 & 0.83 & 2.98 & 0.83 & -7.30 & \textless .00001 \\
    16 & mountain, bushes & 27 & 100.00 & 25.42 & 2.33 & 0.67 & 2.17 & 0.00 & -6.72 & \textless .00001 \\
    18 & field & 40 & 95.00 & 65.77 & 2.64 & 0.72 & 2.38 & 0.72 & -5.83 & \textless .00001 \\
    19 & closet, air\_conditioning & 67 & 89.55 & 68.91 & 1.95 & 1.10 & 1.62 & 1.10 & -3.33 & .000439 \\
    29 & potty, flusher & 76 & 86.84 & 76.64 & 3.83 & 1.73 & 3.62 & 1.73 & -5.03 & \textless .00001 \\
    43 & mountain & 40 & 100.00 & 89.46 & 5.26 & 3.55 & 5.03 & 3.55 & -4.97 & \textless .00001 \\
    \hline
    48 & road & 42 & 100.00 & 77.91 & 6.15 & 1.98 & 6.65 & 1.98 & -8.91 & \textless .00001 \\
    50 & pillow & 42 & 97.62 & 71.36 & 4.43 & 1.08 & 4.47 & 1.08 & -8.57 & \textless .00001 \\
    51 & road, automobile & 84 & 95.24 & 42.41 & 4.15 & 0.00 & 3.79 & 0.00 & -12.78 & \textless .00001 \\
    54 & hedgerow, hedge & 63 & 90.48 & 68.34 & 1.63 & 0.83 & 1.77 & 0.83 & -3.51 & .000222 \\
    62 & buffet & 15 & 93.33 & 31.85 & 1.67 & 0.00 & 1.65 & 0.00 & -5.37 & \textless .00001 \\
    63 & skyscraper & 39 & 94.87 & 52.86 & 2.94 & 0.12 & 2.92 & 0.12 & -8.48 & \textless .00001\\
    \hline
\end{tabular}
\caption{Evaluation details as discussed in Section~\ref{subsubsec:top-3} for third ECII responses.
Images: number of images used for evaluation.
\# Activations: (targ(et)): Percentage of target images activating the neuron (i.e., activation at least 80\% of this neuron's activation maximum); (non-t(arget)): Same for all other images used in the evaluation. Mean/Median (targ(et)/non-t(arget)): Mean/median activation value for target and non-target images, respectively.}
\label{tab:top3_evaluation}
\end{table*}

\subsection{Earlier exploratory data}
\label{subsec:earlier-exploratory}

In this final part of the Appendix, we disclose some exploratory data that we obtained earlier and that informed our study. The data is based on an incompletely trained ResNet50V2 CNN that we later realized did not perform well as a classifier. The data obtained from the analysis of hidden node activations still informed the hypotheses and methods used and reported in the main body of this paper. Since our explorations were deliberately preliminary, they were not carried out with full rigor, and thus have to be interpreted very carefully. We believe that the data, despite these cautions, is still informative and may help in developing follow-up research. 

The data reported is based on modifying certain parameters that go into our analysis, in particular cut-off points for selection of positive and negative examples for Concept Induction analysis, and naively using word lists of the first 10 or even 20 ECII responses as label hypotheses.

\subsubsection{First 10 or 20 ECII Responses}
\label{subsubsec:top-10andtop-20}

In this earlier exploratory investigation
we used label hypotheses consisting of the list of 10 (top-10) or 20 (top-20) ECII responses.

Target images for each label hypothesis were obtained by querying Google Images with the full list. We also used an image augmentation generator which introduced variations to the images before they were fed into the trained model to obtain neuron activations for target labels. In this case, we considered a neuron activated 
if it attained at least 50\% of its maximum activation value.

Table~\ref{tab:top20_and_top10} shows each neuron's target activation percentages for the top-10 and top-20 ECII response label hypotheses. The key insight
is that there is little difference between the top-10 and the top-20 selection, which encouraged us to only look at the first few ECII responses, as we did in the work reported in the main paper.

\begin{table}[t]
\centering
\resizebox{.49\columnwidth}{!}{
\begin{tabular}{c||c|c}
    \multicolumn{1}{l}{Neuron \#} & \multicolumn{2}{c}{Target (\%)} \\
     \hline
     &  top-20 & top-10 \\
    \hline
        0 & 13.58 & 14.81 \\
        \textbf{2} & \textbf{99.63} & \textbf{99.73} \\
        3 & 0.00 & 0.00 \\
        \textbf{4} & \textbf{100.00} & \textbf{100.00} \\
        5 & 40.02 & 39.12 \\
        \hline
        6 & 0.00 & 0.10 \\
        7 & 5.26 & 7.42 \\
        \textbf{8} & \textbf{100.00} & \textbf{100.00} \\
        \textbf{9} & \textbf{99.91} & \textbf{100.00} \\
        10 & 33.33 & 34.86 \\
        \hline
        \textbf{11} & \textbf{99.01} & \textbf{99.04} \\
        \textbf{12} & \textbf{99.01} & \textbf{95.33} \\
        13 & 0.00 & 0.10 \\
        14 & 47.92 & 45.66 \\
        \textbf{15} & \textbf{100.00} & \textbf{100.00} \\
        \hline
        \textbf{16} & \textbf{99.94} & \textbf{100.00} \\
        \textbf{18} & \textbf{98.49} & \textbf{99.89} \\
        \textbf{20} & \textbf{100.00} & \textbf{100.00} \\
        22 & 32.27 & 32.56 \\
        \textbf{23} & \textbf{100.00} & \textbf{100.00} \\
        \hline
        24 & 0.00 & 0.00 \\
        \textbf{25} & \textbf{98.99} & \textbf{98.78} \\
        26 & 0.20 & 0.13 \\
        \textbf{27} & \textbf{99.71} & \textbf{99.78} \\
        29 & 0.47 & 0.61 \\
        \hline
        30 & 0.00 & 0.00 \\
        31 & 0.00 & 0.00 \\
        32 & 0.00 & 0.00 \\
        33 & 0.15 & 0.08 \\
        34 & 57.33 & 63.43 \\
        \hline
        35 & 14.65 & 14.67 \\
        36 & 0.00 & 0.00 \\
        37 & 14.65 & 0.00 \\
        39 & 0.25 & 0.27 \\
        40 & 9.39 & 8.33 \\
        \hline
        \textbf{41} & \textbf{100.00} & \textbf{100.00} \\
        \textbf{42} & \textbf{90.20} & \textbf{88.61} \\
        43 & 0.05 & 0.00 \\
        44 & 4.51 & 1.89 \\
        45 & 0.00 & 0.00 \\
        \hline
        46 & 0.00 & 0.88 \\
        47 & 3.52 & 5.53 \\
        \textbf{48} & \textbf{92.62} & \textbf{92.58} \\
        50 & 0.00 & 0.00 \\
        52 & 0.17 & 0.21 \\
        \hline
        53 & 0.00 & 0.00 \\
        54 & 0.00 & 0.00 \\
        55 & 57.57 & 59.68 \\
        56 & 2.62 & 2.00 \\
        58 & 1.15 & 0.81 \\
        \hline
        59 & 0.00 & 0.00 \\
        \textbf{60} & \textbf{100.00} & \textbf{100.00} \\
        \textbf{62} & \textbf{100.00} & \textbf{100.00} \\
        \textbf{63} & \textbf{100.00} & \textbf{100.00} \\
    \hline
\end{tabular}
}
\caption{Each neuron's activation percentage for top-20 and top-10 label hypothesis. 
The omitted neurons were not activated
by any image, i.e., their maximum activation value was 0. Target \%: Percentage of target images activating the neuron above 50\% of its maximum activation. \textbf{Bold} denotes the 19 neurons whose labels would be considered confirmed.}
\label{tab:top20_and_top10}
\end{table}

\subsubsection{Using different cut-off values}
\label{subsubsec:cutoffs}

In this early exploratory investigation we looked at different cut-off values for selecting the positive and negative image sets that go into the Concept Induction analysis to generate label hypotheses for the neurons. These cut-offs provide valuable insights into the impact of activation thresholds on neuron behavior and target label recognition. 

We looked at the following cases:
\begin{enumerate}
\item The positive set consists of images that exhibit an activation of at least 50\% of the maximum activation value, while the negative set consists of images with activation below 50\% of the highest activation value.\label{case1}
\item The positive set consists of images with activations of at least 50\% of the maximum activation value. The negative set consists of images not activating the neuron at all (i.e., activation value exactly zero).\label{case2}
\item The positive set consists of images with activations strictly greater than zero, i.e., activation values between 0\% and the maximum activation value. The negative set consists of images not activating the neuron at all (i.e., the activation value is exactly zero).\label{case3}

\item The positive set consists of images with activations of at least 80\% of the maximum activation value, while the negative set consists of images with activations below 80\%.\label{case4}
\end{enumerate}

To generate label hypotheses for each neuron in all four cases, we considered the list of top-50 ECII responses (in the way we looked at top-10 and top-20 for the data reported above) sorted by coverage score. 
Table~\ref{tab:top50_allCases} shows the activation percentages of each neuron for the target labels in the different cut-off cases. 
The results indicate that there is little difference between the cases. This encouraged us to use more selective cut-off values for the reported study, with positive examples selected 
from images with activations of at least 80\% of the maximum activation value (as in Case~\ref{case4}) while negative examples were chosen as images activating the neuron with at most 20\% of the maximum activation value.

\begin{table}[t]
\centering
\begin{footnotesize}
\begin{tabular}{c||c|c|c|c}
    Neuron \# & Case~\ref{case1} & Case~\ref{case2} & Case~\ref{case3} & Case~\ref{case4} \\
    \hline
        0 & 15.47 & 12.78 & 13.65 & 17.28 \\
        \textbf{2} & \textbf{99.37} & \textbf{99.39} & \textbf{99.09} & \textbf{99.44} \\
        3 & 0.00 & 0.00 & 0.00 & 0.00 \\
        \textbf{4} & \textbf{100.00} & \textbf{100.00} & \textbf{100.00} & \textbf{100.00} \\
        5 & 35.01 & 36.84 & 38.38 & 45.17 \\
        \hline
        6 & 0.44 & 0.40 & 0.24 & 0.22 \\
        7 & 6.31 & 7.48 & 5.71 & 5.88 \\
        \textbf{8} & \textbf{100.00} & \textbf{100.00} & \textbf{100.00} & \textbf{100.00} \\
        \textbf{9} & \textbf{99.90} & \textbf{100.00} & \textbf{99.97} & \textbf{99.80} \\
        10 & 34.40 & 34.33 & 34.43 & 30.17 \\
        \hline
        \textbf{11} & \textbf{99.00} & \textbf{99.00} & \textbf{99.00} & \textbf{99.73} \\
        \textbf{12} & \textbf{95.20} & \textbf{95.20} & \textbf{95.20} & \textbf{96.85} \\
        13 & 0.05 & 0.06 & 0.05 & 0.15 \\
        14 & 50.48 & 47.02 & 47.02 & 38.70 \\
        \textbf{15} & \textbf{99.93} & \textbf{99.96} & \textbf{100.00} & \textbf{99.89} \\
        \hline
        \textbf{16} & \textbf{99.94} & \textbf{99.97} & \textbf{99.97} & \textbf{100.00} \\
        \textbf{18} & \textbf{99.09} & \textbf{99.39} & \textbf{99.60} & \textbf{99.58} \\
        \textbf{20} & \textbf{100.00} & \textbf{100.00} & \textbf{100.00} & \textbf{100.00} \\
        22 & 37.32 & 26.00 & 26.24 & 35.62 \\
        \textbf{23} & \textbf{100.00} & \textbf{100.00} & \textbf{100.00} & \textbf{99.90} \\
        \hline
        24 & 0.05 & 0.05 & 0.06 & 0.09 \\
        \textbf{25} & \textbf{98.61} & \textbf{98.34} & \textbf{98.02} & \textbf{97.11} \\
        26 & 0.23 & 0.17 & 0.23 & 0.53 \\
        \textbf{27} & \textbf{99.64} & \textbf{99.65} & \textbf{99.60} & \textbf{99.25} \\
        29 & 0.67 & 0.67 & 0.67 & 0.10 \\
        \hline
        30 & 0.00 & 0.06 & 0.08 & 0.00 \\
        31 & 0.00 & 0.00 & 0.00 & 0.00 \\
        32 & 0.00 & 0.00 & 0.00 & 0.00 \\
        33 & 0.19 & 0.19 & 0.22 & 0.19 \\
        34 & 56.46 & 56.46 & 57.58 & 58.44 \\
        \hline
        35 & 16.32 & 16.31 & 9.25 & 5.12 \\
        36 & 0.00 & 0.00 & 0.00 & 0.00 \\
        37 & 0.00 & 0.00 & 0.00 & 0.06 \\
        39 & 0.24 & 0.20 & 0.63 & 0.22 \\
        40 & 8.39 & 8.91 & 11.12 & 10.07 \\
        \hline
        \textbf{41} & \textbf{100.00} & \textbf{99.95} & \textbf{100.00} & \textbf{99.95} \\
        \textbf{42} & \textbf{91.07} & \textbf{90.71} & \textbf{89.56} & \textbf{87.88} \\
        43 & 0.03 & 0.10 & 0.04 & 0.00 \\
        44 & 6.35 & 6.71 & 6.92 & 6.78 \\
        45 & 0.00 & 0.09 & 0.00 & 0.00 \\
        \hline
        46 & 0.91 & 0.43 & 0.21 & 0.14 \\
        47 & 3.12 & 4.55 & 1.62 & 2.65 \\
        \textbf{48} & \textbf{91.68} & \textbf{93.01} & \textbf{92.43} & \textbf{90.28} \\
        50 & 0.00 & 0.00 & 0.00 & 0.00 \\
        52 & 0.18 & 0.24 & 0.17 & 0.04 \\
        \hline
        53 & 0.00 & 0.00 & 0.00 & 0.00 \\
        54 & 0.00 & 0.00 & 0.00 & 0.00 \\
        55 & 53.81 & 47.88 & 38.36 & 33.55 \\
        56 & 4.16 & 4.06 & 2.22 & 2.75 \\
        58 & 1.80 & 1.80 & 1.68 & 0.45 \\
        \hline
        59 & 0.00 & 0.00 & 0.00 & 0.13 \\
        \textbf{60} & \textbf{100.00} & \textbf{99.96} & \textbf{100.00} & \textbf{99.80} \\
        \textbf{62} & \textbf{100.00} & \textbf{100.00} & \textbf{100.00} & \textbf{100.00} \\
        \textbf{63} & \textbf{100.00} & \textbf{100.00} & \textbf{100.00} & \textbf{99.94}\\
    \hline
\end{tabular}
\end{footnotesize}
\caption{Activation percentage for top-50 label hypothesis for each neuron.
The omitted neurons were not activated
by any image, i.e., their maximum activation value was 0. The four cases have different cut-offs as described in the text. \textbf{Bold} denotes the 19 neurons whose labels would be considered confirmed.}
\label{tab:top50_allCases}
\end{table}